\documentclass[12pt]{spieman}  
\usepackage{amsmath,amsfonts,amssymb}
\usepackage{graphicx}
\usepackage{setspace}
\usepackage{tocloft}

\usepackage{epsfig, subfig}
\usepackage{float}
\usepackage{times}
\usepackage{color}

\title{Classifying Symmetrical Differences and Temporal Change for the Detection of Malignant Masses in Mammography Using Deep Neural Networks}

\author[a, *]{Thijs Kooi}
\author[a]{Nico Karssemeijer}
\affil[a]{Diagnostic Image Analysis Group, Department of Radiology and Nuclear Medicine, RadboudUMC Nijmegen, The Netherlands}

\cftpagenumbersoff{figure}
\cftpagenumbersoff{table} 
\begin{document} 
\maketitle

\begin{abstract}
  We investigate the addition of symmetry and temporal context information to a deep convolutional neural network (CNN) with the purpose of detecting malignant soft tissue lesions in mammography. We employ a simple linear mapping that takes the location of a mass candidate and maps it to either the contra-lateral or prior mammogram and regions of interest (ROI) are extracted around each location. Two different architectures are subsequently explored: (1) a fusion model employing two datastreams were both ROIs are fed to the network during training and testing and (2) a stage-wise approach where a single ROI CNN is trained on the primary image and subsequently used as feature extractor for both primary and contra-lateral or prior ROIs. A 'shallow' gradient boosted tree (GBT) classifier is then trained on the concatenation of these features and used to classify the joint representation. \\
  
  The baseline obtained an AUC of $0.87$ with confidence interval $[0.853, 0.893]$. For the analysis of symmetrical differences, the first architecture where both primary and contra-lateral patches are presented during training obtained an AUC of $0.895$ with confidence interval $[0.877, 0.913]$ and the second architecture where a new classifier is retrained on the concatenation an AUC of $0.88$ with confidence interval $[0.859, 0.9]$. We found a significant difference between the first architecture and the baseline at high specificity $p = 0.02$. When using the same architectures to analyze temporal change we obtained an AUC of $0.884$ with confidence interval $[0.865, 0.902]$ for the first architecture and an AUC of $0.879$ with confidence interval $[0.858, 0.898]$ in the second setting. Although improvements for temporal analysis were consistent, they were not found to be significant. The results show our proposed method is promising and we suspect performance can greatly be improved when more temporal data becomes available.
\end{abstract}


{\noindent \footnotesize\textbf{*}Corresponding author,  \linkable{email@thijskooi.com} }

\section{Introduction}
\label{sec::introduction}
Breast cancer screening in the form of annual or biennial breast X-rays is being performed to detect cancer at an early stage. This has been shown to increase chances of survival significantly, with some studies showing a reduction in breast cancer mortality of up to 40\% \cite{Taba03}. Human reading of screening data is time consuming and error prone and to aid interpretation, computer aided detection and diagnosis (CAD) \cite{Gige01, Doi07, Doi05, Ginn11} systems are developed. For mammography, CAD is already widely applied as a second reader \cite{Rao10, Mali06} but the effectiveness of current technology is disputed. Several studies show no increase in sensitivity or specificity with CAD \cite{Tayl05b} for masses or even a decreased specificity without an improvement in detection rate or characterization of invasive cancers \cite{Fent11, Lehm15}. \\ 

During a mammographic exam, images are typically recorded of each breast and absence of a certain structure around the same location in the contra-lateral image will render an area under scrutiny more suspicious. Conversely, the presence of a similar tissue less so. Additionally, due to the annual or biennial organization of screening, there is a temporal dimension and similar principles apply: the amount of tissue is expected to decrease, rather than increase with age and therefore, novel structures that are not visible on previous exams, commonly referred to as {\it priors}, spark suspicion. \\

In medical literature, an asymmetry denotes a potentially malignant density that is not characterized as a mass or architectural distortion. Four types are distinguished: (1) a plain {\it asymmetry} refers to a density lacking convex borders, seen in only one of the two standard mammographic views, (2) a {\it focal asymmetry} is visible on two views but does not fit the definition of a mass, (3) a {\it global asymmetry} indicates a substantial difference in total fibroglandular tissue between left and right breast, (4) a {\it developing asymmetry} refers to a growing asymmetry in comparison to prior mammograms \cite{Sick07, Youk09}. These types are generally benign, but have been associated with an increased risk \cite{Scut06} and are sometimes the only manifestation of a malignancy. To the best of our knowledge, no relevant work has been done that compares reader performance of malignancies with and without left and right comparisons, but asymmetry is often mentioned by clinicians as an important clue, also to detect malignancies that are classified as a mass. The merit of temporal comparison mammograms on the other hand has been well studied and is generally known to improve specificity without a profound impact on sensitivity for detection \cite{Thur00, Burn02, Vare05, Roel07, Yank11}. \\

Burnside et al. \cite{Burn02} analyzed a set of diagnostic and screening mammograms and concluded that in the latter case, comparison with previous examinations significantly decreases the recall rate and false positive rate, but does not increase sensitivity. Varela et al. \cite{Vare05} compared the reading performance of six readers and found the performance drops significantly when removing the prior mammogram, in particular in areas of high specificity, relevant for screening. Roelofs et al. \cite{Roel07} also investigated the merit of prior mammograms in both detection and assessment of malignant lesions. Their results show performance was significantly better in the presence of a prior exam, but no more lesions were found. They subsequently postulate priors are predominantly useful for interpretation and less so for initial detection. Yankakis et al. \cite{Yank11} additionally investigate the effect of noticeable change in tissue in mammograms. They generated separate sets of current-prior examination pairs with and without noticeable change and observed that recall rate, sensitivity and cancer detection rate (CDR) are higher when change is noted, but specificity is lower, resulting in a higher false positive rate. \\

Symmetry is often used as a feature in traditional CAD systems detecting pathologies such as lesions in the brain \cite{Lui09h}, prostate cancer \cite{Litj14c} and abnormalities in the lungs \cite{Ginn01a}. Most research on mammographic asymmetries involves the classification of a holistic notion of discrepancy rather than the incorporation of this information in a CAD system \cite{Ferr01, Cast15}. Published work on temporal analysis typically relies on the extraction of features from both current and prior exams which are combined into a single observation and fed to a statistical learning algorithm \cite{Hadj01a, Timp07}. For detection, an additional registration step is performed \cite{Timp06}. This has been shown to significantly increase performance of the traditional, handcrafted feature based systems. \\

Recent advances in machine learning, in particular deep learning \cite{Lecu98, Hint06a, Beng13, Schm15a} signified a breakthrough in artificial intelligence (AI) and several pattern recognition applications are now claiming human or even super human performance \cite{Cire12a, Mnih15, Ioff15, Silv16}. Deep convolutional neural networks (CNNs) \cite{Lecu98} are currently dominating leader boards in challenges for both natural \cite{Russ14a} and medical image analysis challenges \cite{Cire13a, Ronn15, Wang16a}. Rather than relying on engineers and domain experts to design features, the systems learn feature transformations from data, saving enormous amounts of time in development. The adoption of deep neural networks in medical image analysis was initially reluctant, but the community has recently seen a surge of papers \cite{Litj17} some showing significant improvements upon the state-of-the art \cite{Roth16, Kooi16, Seti16, Grin16b}. \\

The vanilla CNN architecture is a generic problem solver for many signal processing tasks but is still limited by the constraint that a single tensor needs to be fed to the front-end layer, if no further adaptations to the network are made. Medical images provide an interesting new data source, warranting adaptation of methods successful in natural images. Several alternative architectures that go beyond the patch level and work with multi-scale \cite{Fara13a} or video \cite{Karp14, Neve14, Simo14a} have been explored for natural scenes. In these settings, multiple datastreams are employed, where each datastream represents, for instance, a different scale in the image or frames at different time points in a video. Similar ideas have been applied to medical data, most notably the 2.5D simplification of volumetric scans \cite{Pras13a, Roth14, Roth16}. \\

\begin{figure}
  \centering
  \subfloat[Primary]{\includegraphics[width=0.15\textwidth]{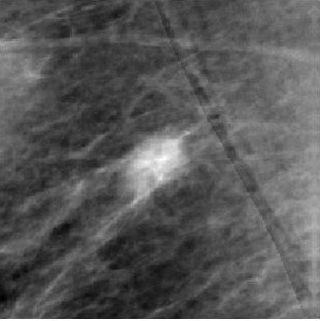}}\hspace{0.5cm}
  \subfloat[Secondary]{\includegraphics[width=0.15\textwidth]{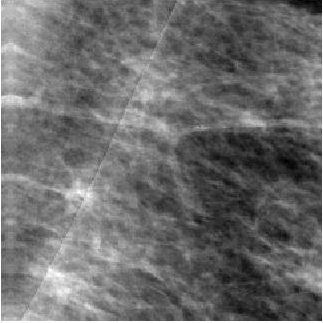}}

  \subfloat[Primary]{\includegraphics[width=0.15\textwidth]{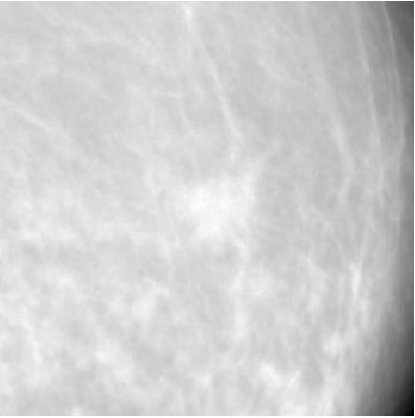}}\hspace{0.5cm}
  \subfloat[Secondary]{\includegraphics[width=0.1503\textwidth]{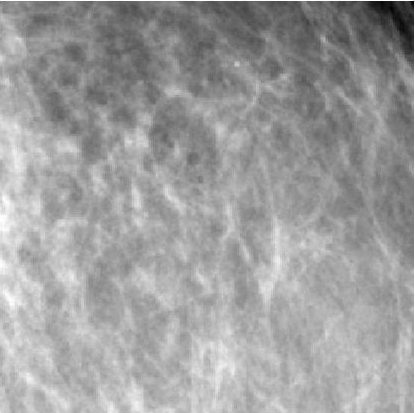}}
  
  \subfloat[Primary]{\includegraphics[width=0.15\textwidth]{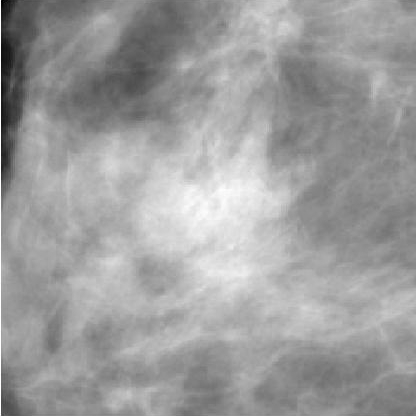}}\hspace{0.5cm}
  \subfloat[Secondary]{\includegraphics[width=0.15\textwidth]{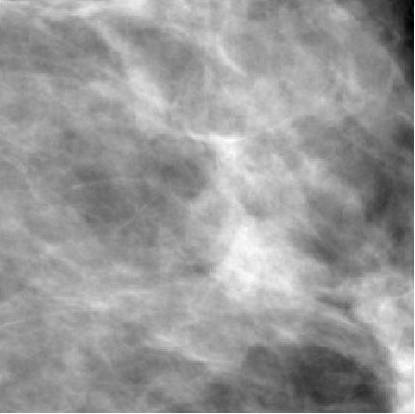}}

  \caption{Examples of symmetry pairs. {\it Top row:} Very suspicious malignant lesion, regardless of its contra-lateral counterpart. \\
  {\it Middle row:} Malignant lesion that is more suspicious in the light of its contra-lateral image. \\
  {\it Bottom row:} Normal structure that is less suspicious in the light of its contra-lateral image. }
  \label{fig::symmetry_pair_examples}
\end{figure}

\begin{figure}
 \centering
 \includegraphics[width=0.2\textwidth]{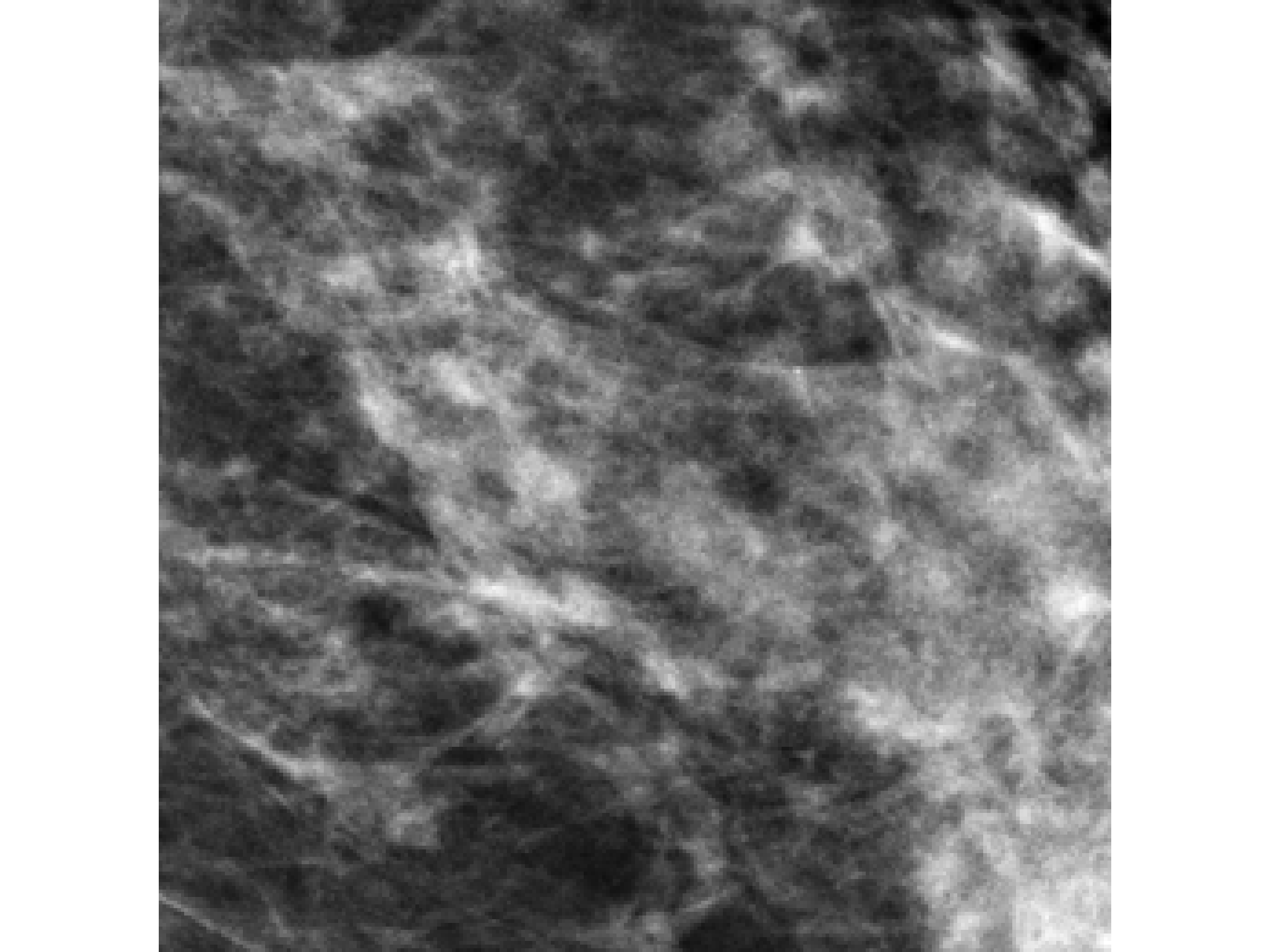}
 \includegraphics[width=0.2\textwidth]{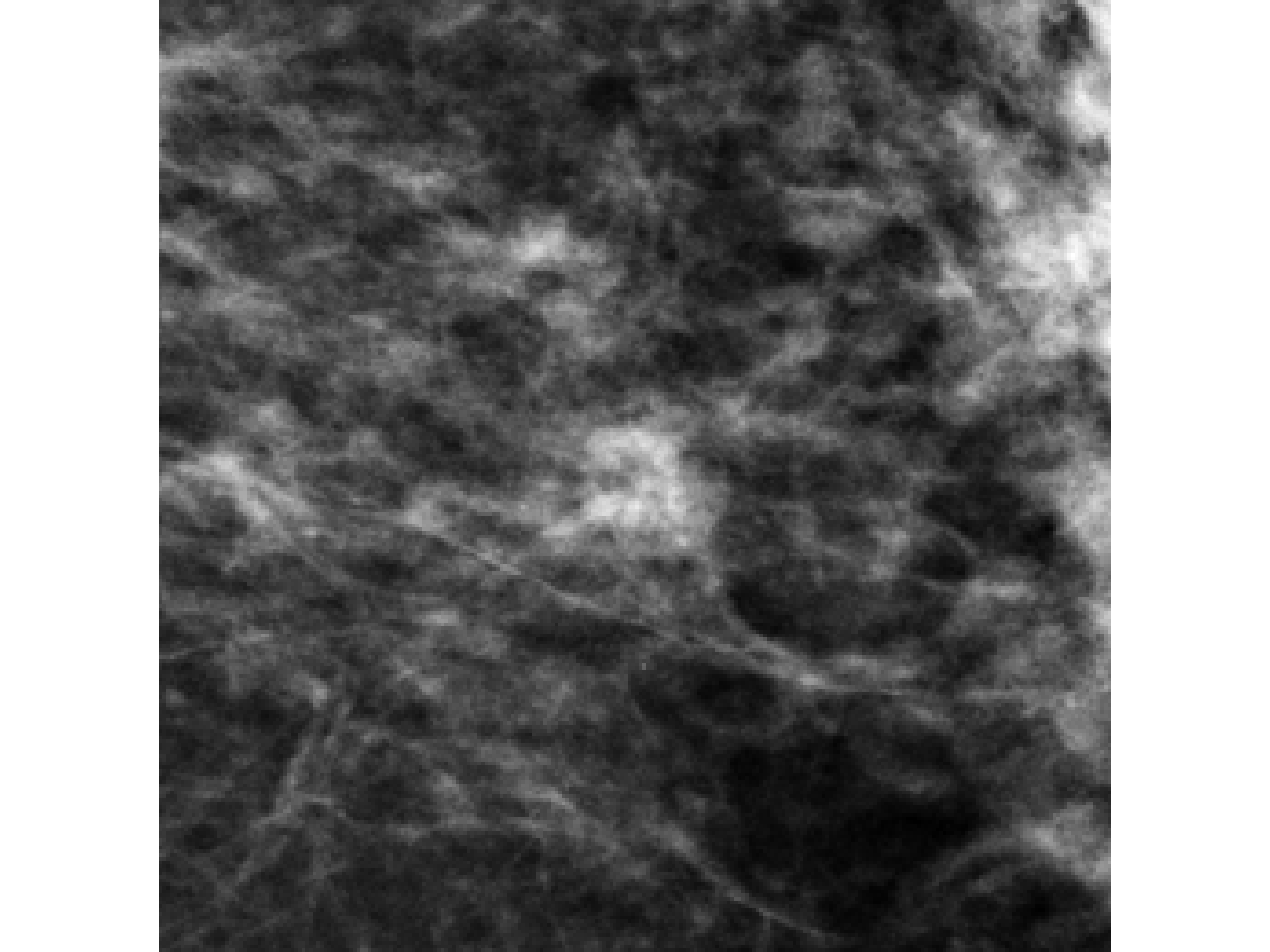} \\\vspace{0.4cm}
 
 \includegraphics[width=0.2\textwidth]{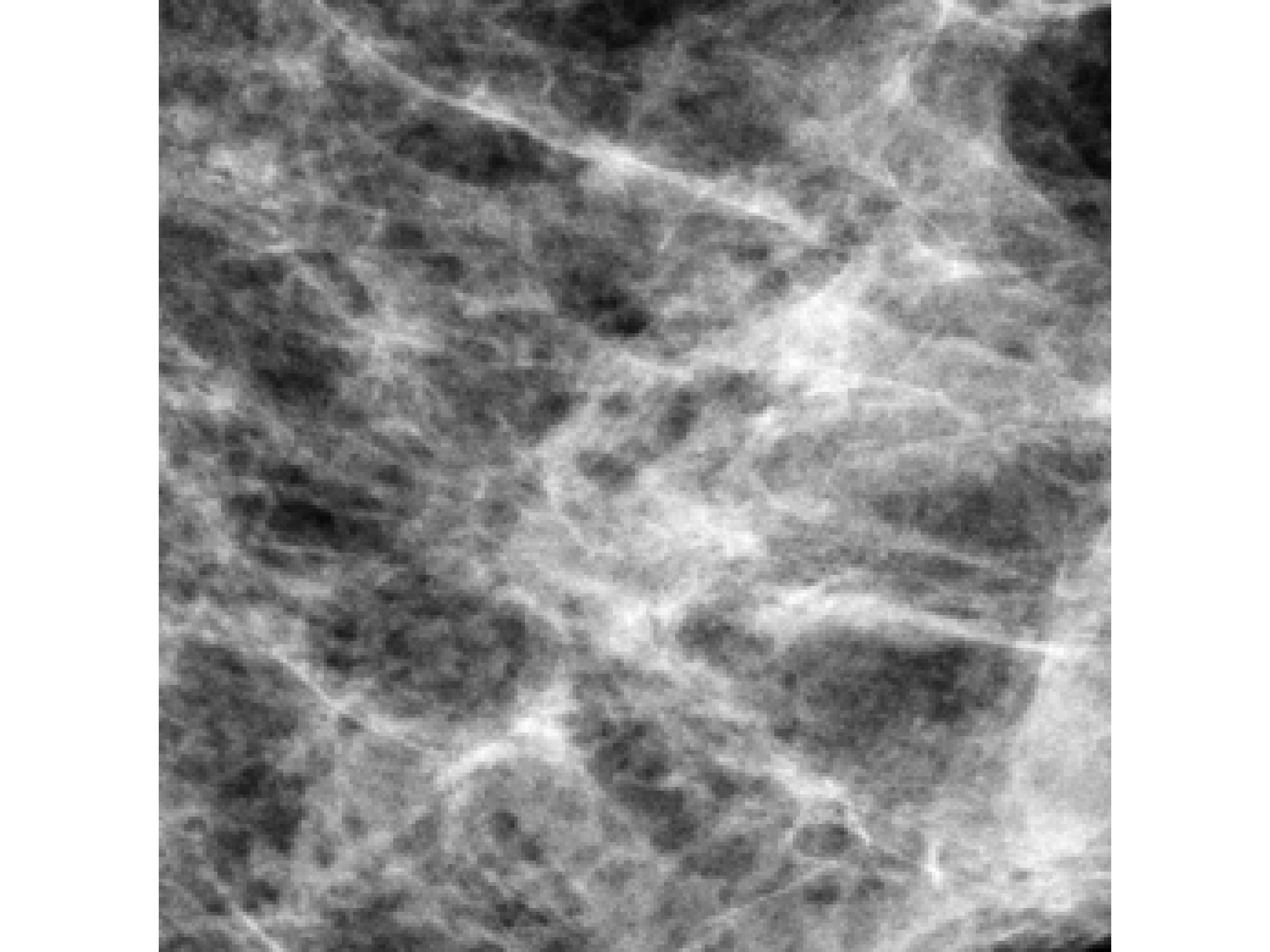}
 \includegraphics[width=0.2\textwidth]{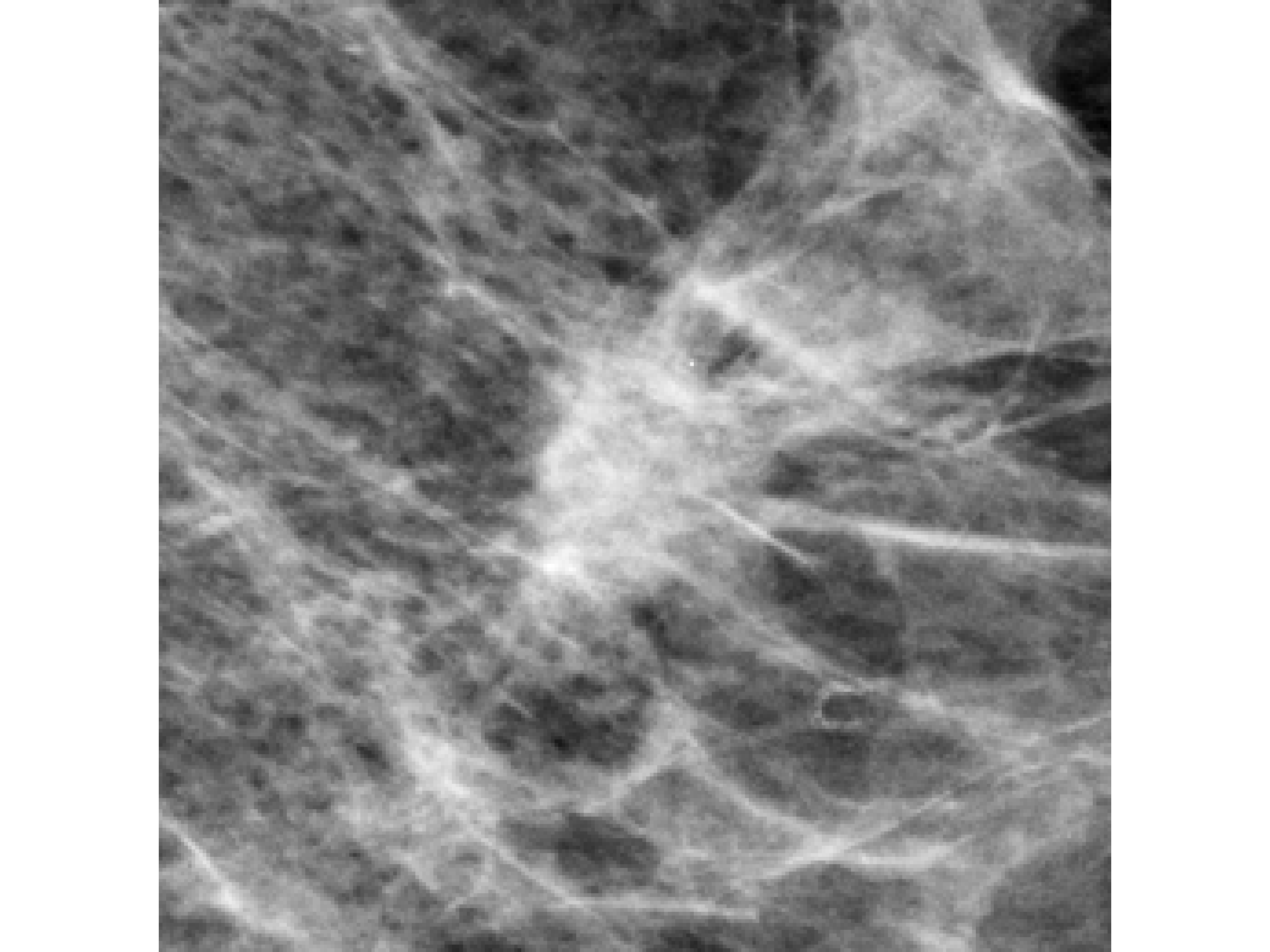} \\\vspace{0.4cm}
 
  \includegraphics[width=0.2\textwidth]{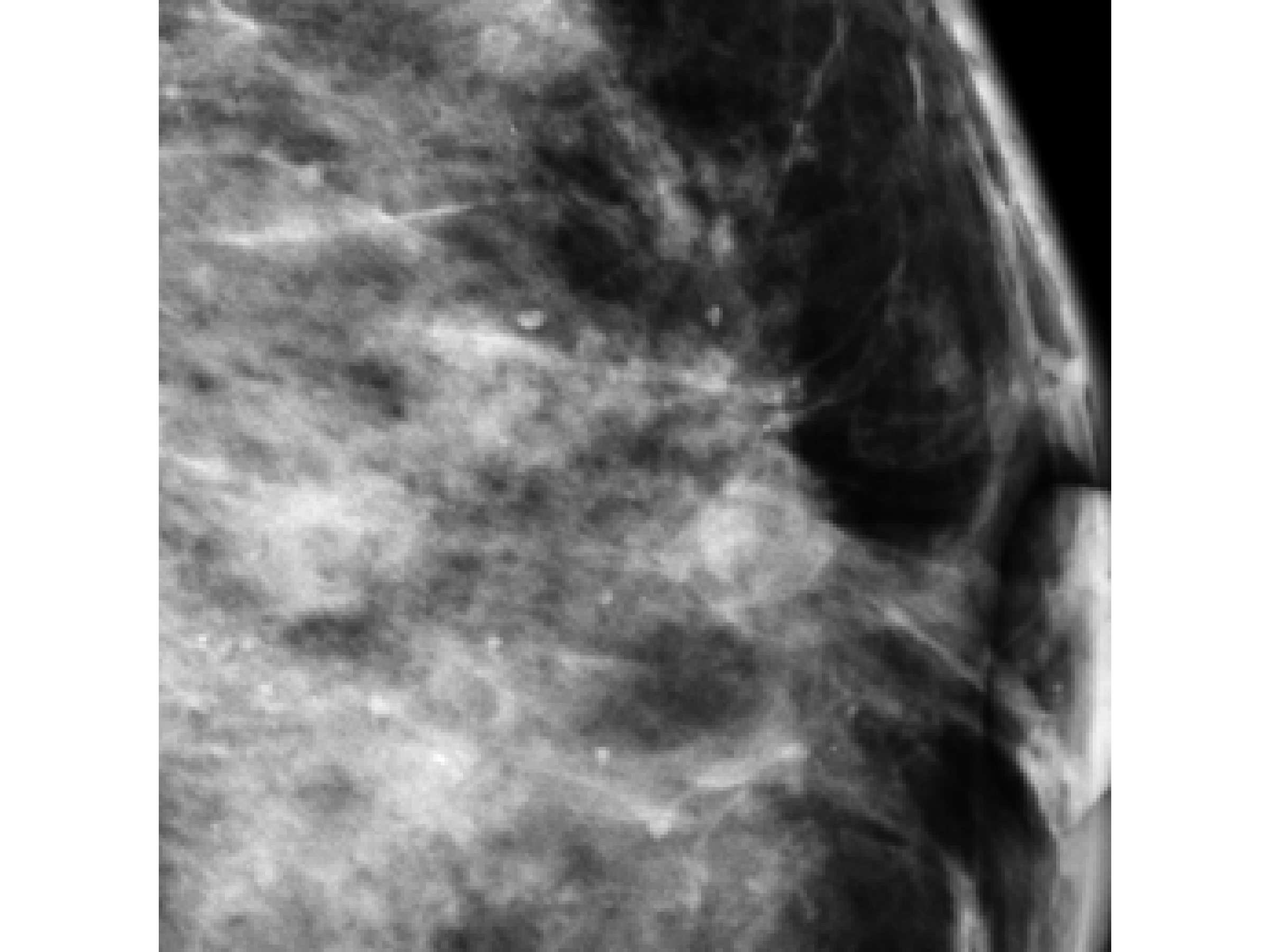}
 \includegraphics[width=0.2\textwidth]{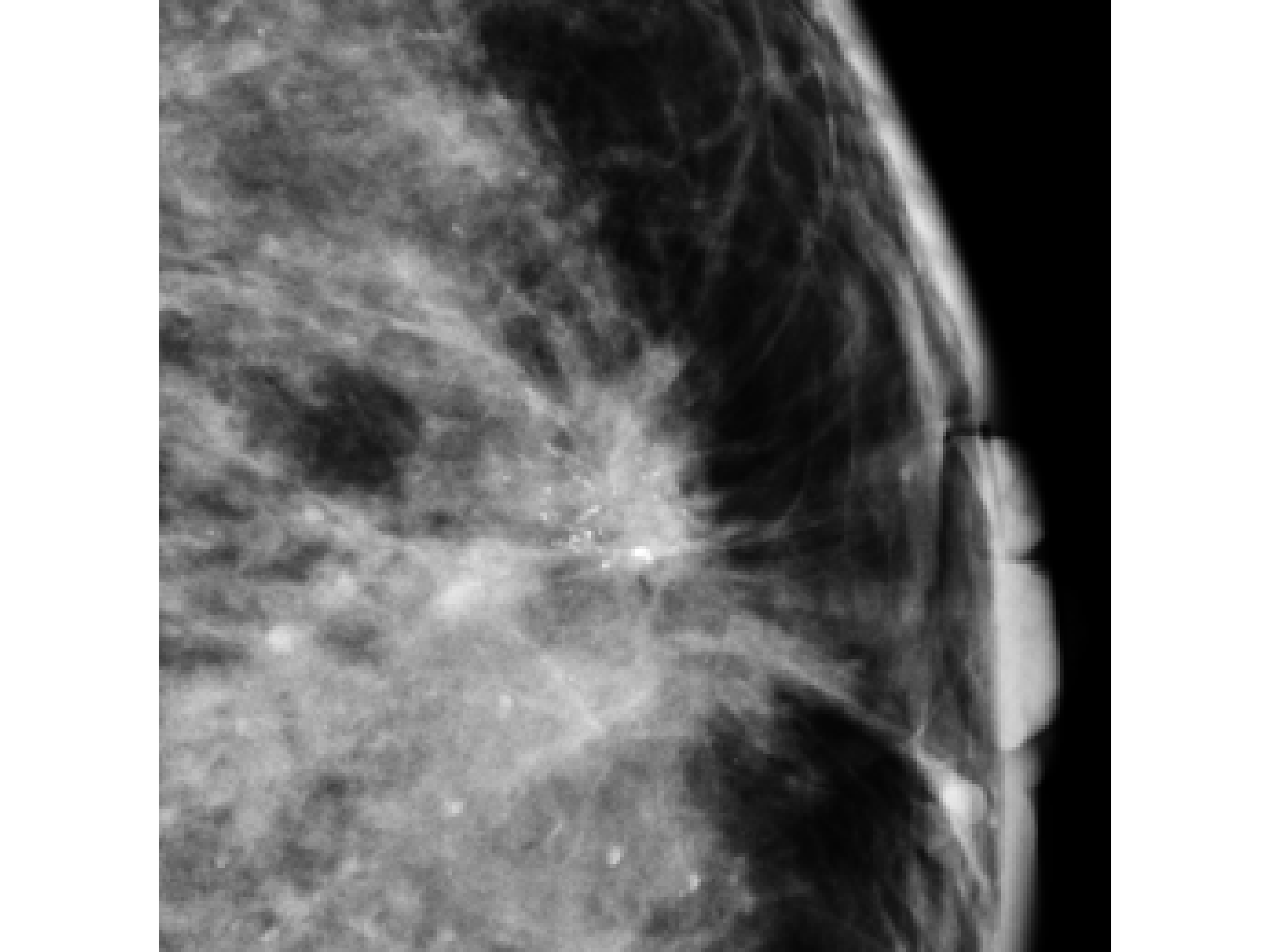} \\
 \caption{Examples of temporal pairs. The right column represents the current and the left column the prior image it is compared with, using the mapping described in section \ref{sec::mapping_locations}}
 \label{fig::temporal_pair_examples}
\end{figure}

In this paper we extend previous work \cite{Kooi17b} and investigate the addition of symmetry and temporal information to a deep CNN with the purpose of detecting malignant soft tissue lesions in mammography. We employ a simple linear mapping that takes the location of a mass candidate and maps it to either the contra-lateral or prior mammogram and regions of interest (ROI) are extracted around each location. We subsequently explore two different architectures
\begin{enumerate}
 \item A fusion model employing two datastreams were both ROIs are fed to the network during training and testing.
 \item A stage-wise approach where a single ROI CNN is trained on the primary image and subsequently used as feature extractor for both primary and contra-lateral or prior ROIs. A 'shallow' gradient boosted tree (GBT) classifier is subsequently trained on the concatenation of these features and used to classify similar concatenations of features in the test set. 
\end{enumerate}
Examples of symmetry pairs are show in figure \ref{fig::symmetry_pair_examples}. Figure \ref{fig::temporal_pair_examples} shows several examples of temporal pairs. \\

To the best of our knowledge, this is the first CAD and deep learning approach incorporating symmetry as a feature in a CAD system and the first CAD paper exploring deep neural networks for temporal comparison. Even though the methods are applied to mammography, we feel results may be relevant as well for other medical image analysis tasks, where classification of anomalies that occur unilaterally or develop over time is important, such as lung, prostate and brain images. \\

The rest of this paper is divided into 5 sections. In the following section, we will outline the data pre-processing, candidate detector and linear mapping used. In section 3 the deep neural architectures will be described followed by a description of the data and experimental setup in section 4. Results will be discussed in section 5 and we will end with a conclusion in section 6.

\section{Methods}
\subsection{Candidate Detection}
\label{sec::candidate_detection}
We generally follow the candidate detection setup described in Kooi et al. \cite{Kooi16}. To get potential locations of lesions and extract candidate patches, we make use of a popular candidate detector for mammographic lesions \cite{Kars96a}. It employs five features based on first and second order Gaussian kernels, two designed to spot the center of a focal mass and two looking for spiculation patterns, characteristic of malignant lesions. A final feature indicates the size of optimal response in scale-space. We subsequently apply a random forest \cite{Brei01} classifier to generate a likelihood map on which we perform non-maximum suppression. All optima are treated as candidates and patches of $250 \times 250$ pixels, or 5 cm at 200 micron, are extracted around each center location. Since many candidates are too close to the border to extract full patches, we pad the image with zeros. \\

For data augmentation, we follow the scheme described in Kooi et al. \cite{Kooi16}. Each patch in the training set containing an annotated malignant lesion is translated 16 times by adding values sampled uniformly from the interval $[-25, 25]$ ($0.5$ cm) to the lesion center. Each original positive patch is scaled 16 times by adding values sampled uniformly from the interval $[-30, 30]$ ($0.6$ cm) to the top left and bottom right of the bounding box. All patches, both positive and negative are rotated using four 90 degree rotations. This results in $(1+16+16)4 = 132$ patches per positive lesions and $4$ per negative. In practice, these operations are computed on the fly during training, to prevent large datasets on disk. After candidates have been generated, locations are mapped to the same point in the contra-lateral image or the prior. 
\subsection{Mapping Image Locations}
\label{sec::mapping_locations}
\begin{figure}
 \centering
 \subfloat[Primary Image]{\includegraphics[width=0.25\textwidth]{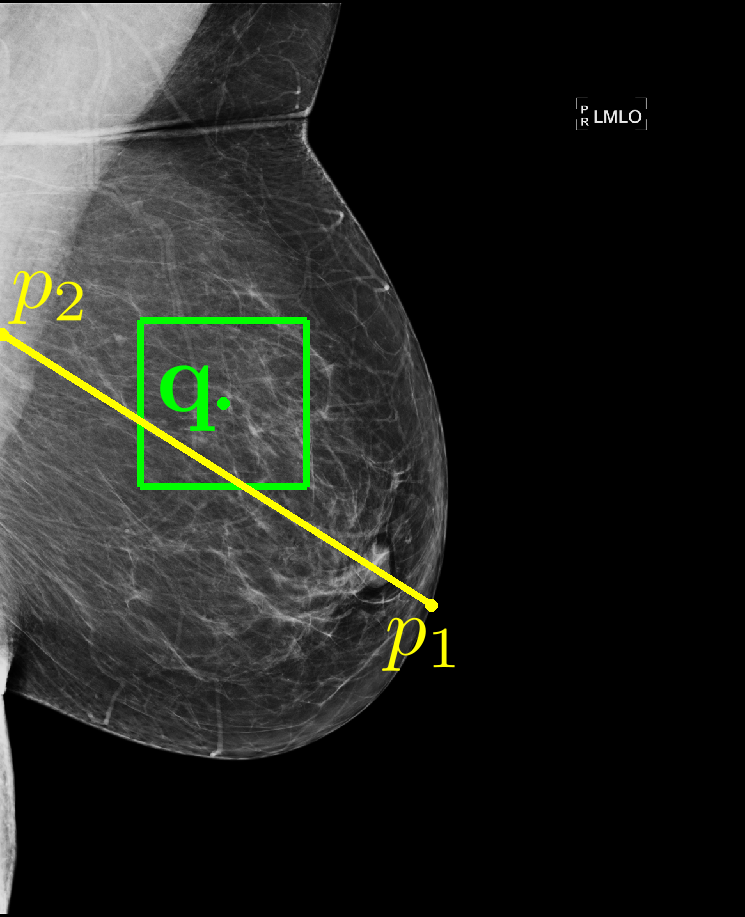}} \\
 \subfloat[Prior Image]{\includegraphics[width=0.2\textwidth]{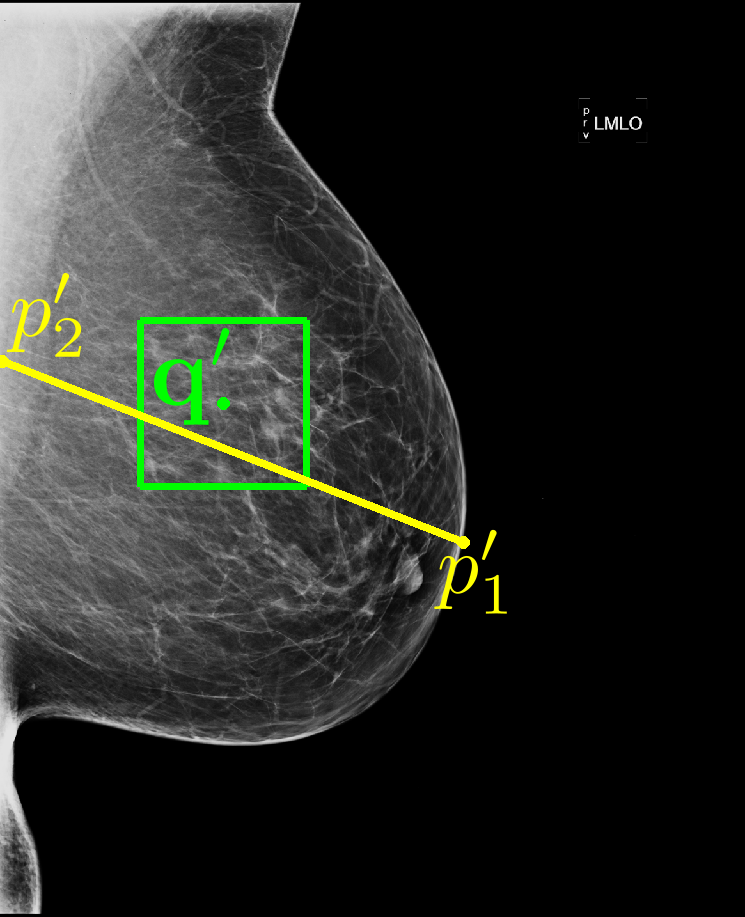}}\hspace{0.5cm}
 \subfloat[Contra-lateral Image]{\includegraphics[width=0.2\textwidth]{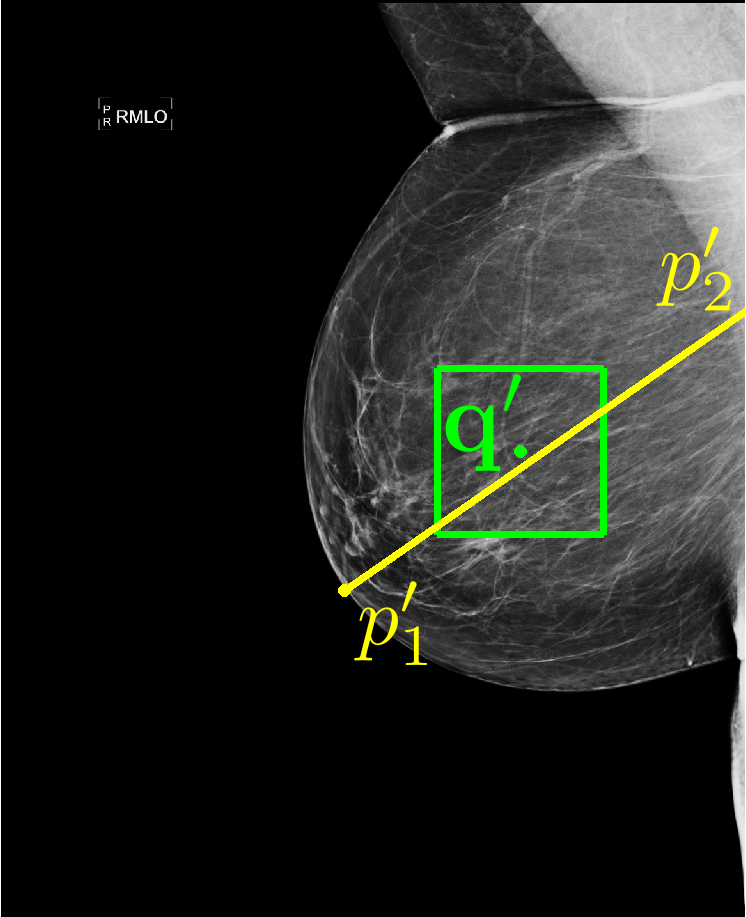}}
 
 \caption{To incorporate symmetry and temporal information, we make use of a simple mapping, based on two coordinates indicated by the end points of the yellow line. (a) A Region Of Interest (ROI) represented by the green box is extracted around a potential malignant lesion location, indicated by the green dot, found by a candidate detector. The location is subsequently matched to either the prior (b) or the contra-lateral image (c). We explore two deep Convolutional Neural Network (CNN) fusion strategies to optimally capture the relation between contra-lateral and prior images.}
 \label{fig::symmetry_mapping_example}
\end{figure}
Finding corresponding locations between two mammograms is a challenging problem due to two main factors: (1) apart from the nipple and chest wall, which may not always be visible, there are no clear landmarks to accommodate feature based registration and (2) the transformation is highly non-linear. Before the mammogram is recorded the breast is deformed strongly: the viewing area is optimized and dose is minimized by stretching the breast. Additionally, the compression plates may not always touch the breast at the same location causing some movement of tissue within the breast. \\

A comparative study between several commonly applied registration methods by Van Engeland et al. \cite{Enge03} found a simple linear approach based on the position of the nipple and center of mass alignment outperformed more complex methods such as warping. We propose a similar approach based on two landmarks. To obtain these points, the whole breast area is first segmented using simple thresholding, followed by a linear hough transform to segment the pectoral muscle \cite{Kars98}, in the case of an MLO image. The row location of the front of the breast (an approximation of the nipple location) $p_{1}$ is subsequently estimated by taking a point on the contour of the breast with the largest distance to the line output by the hough transform. A column point in the pectoral muscle or chest wall $p_{2}$ is taken by drawing a straight line from this point perpendicular to the fit output by the hough transform. The lesion center in the image under evaluation ${\bf q} = (q_{r}, q_{c})^{T}$, where $q_{r}$ and $q_{c}$ denote the row and column location, respectively, is subsequently mapped to the estimated lesion center ${\bf q}'$ in the contra-lateral or prior image according to:
\begin{equation}
 {\bf q'} = {\bf q} - {\bf p} + {\bf p'}
\end{equation}
with ${\bf p} = (p_{1}, p_{2})^{T}$ and ${\bf p'} = (p'_{1}, p'_{2})^{T}$ the same points in the contra-lateral or prior mammogram. In other words, we simply clamp the x-distance to the chest wall and the y-distance to the estimated location of the nipple. An example is provided in figure \ref{fig::symmetry_mapping_example}.\\

Since most CNN architectures induce a decent amount of translation invariance, the mapping does not need to be very precise. To further mitigate mapping errors, we introduce a form of data augmentation by mapping each location in the image in question to 64 different points in the comparison mammogram by sampling the location from a Gaussian with zero mean and 10 pixel standard deviation. 

\section{Deep Convolutional Neural Networks}
\label{sec::multi_strean_cnns}
The CNN architecture exploits structure of the input by sharing weights at different locations in the image, resulting in a convolution operation, the main workhorse of the CNN. The main difference between deep models and conventional statistical learning methods is the nested non-linear function the architecture represents. At each layer, the input signal is convolved with a set of $K$ kernels $\mathcal{W} = ({\bf W}_{1}, {\bf W}_{2}, \ldots, {\bf W}_{K})$ and biases $\mathcal{B} = \{b_{1}, b_{2}, \ldots, b_{K}\}$ are added, each generating a new set of feature maps ${\bf X}_{k}$. These features are subjected to an element-wise non-linear transform $\sigma(\cdot)$ and the same process is repeated for every convolutional layer $l_{0}, l_{1}, \ldots, l_{L}$:
\begin{equation}
 {\bf X}_{k}^{l} = \sigma\big( {\bf W}_{k}^{l -1} \otimes {\bf X}^{l -1} + b_{k}^{l-1} \big)
\end{equation}
Convolutional layers are generally alternated with pooling layers that subsample the resulting feature maps, generating some translation invariance and reducing the dimensionality as information flows through the architecture. After these layers, the final tensor of feature maps is flattened to a vector ${\bf x}^{l}$ and several fully connected layers are typically added, where weights are no longer shared: 
\begin{equation}
 {\bf x}^{l} = \sigma({\bf W}^{l}{\bf x}^{l-1} + b^{l})
\end{equation}
The posterior distribution over a class variable $y_{i}$, given input patch ${\bf X}^{0}$ is acquired by feeding the last level of activations ${\bf x}^{L}$ to either a logistic sigmoid for single class or a softmax function for multi class:
\begin{equation}
 P(y_{i} | {\bf X}^{0}; \Theta) = \text{softmax}({\bf x}^{L}; {\bf W}, {\bf b}) = \frac{e^{{\bf w}_{i}^{T}{\bf x}^{L} + b^{L}_{i}}}{\sum^{K}_{k = 1} e^{{\bf w}_{k}^{T}{\bf x}^{L} + b^{L}_{k}}}
\end{equation}
with $\Theta$ the set of all weights and biases in the network and ${\bf w}_{i}$ the vectorized set of weights leading to the output node of class $i$. The whole network can be seen as a parameterized feature extractor and classifier, where the parameters of the feature transformation and classifier are learned jointly and optimized based on training data. \\

The parameters in the network are generally learned using Maximum Likelihood Estimation (MLE) or Maximum A-Posteriori (MAP), when employing regularization and default backpropagation. Increasing depth up to some point, seems to improve efficiency and reduces the amount of parameters that need to be learned, without sacrificing performance or even increases overall performance \cite{Sriv15, Simo14, He15b}. The gradient of the error of each training sample is dispersed among parameters in every layer during backpropagation and hence becomes smaller (or in rare cases explodes), which is referred to as the fading gradient problem. Common tricks to quell this phenomenon are smart weight initialization \cite{Glor10, He15}, batch normalization \cite{Ioff15} and non-saturating transfer functions such as Rectified Linear Units (ReLU) or recently Exponential Linear Units (ELU) \cite{Clev15}. 

\subsection{Fusion Architectures}
\label{sec::fusion_architectures}
\begin{figure}
 \centering
 \includegraphics[width=0.65\textwidth]{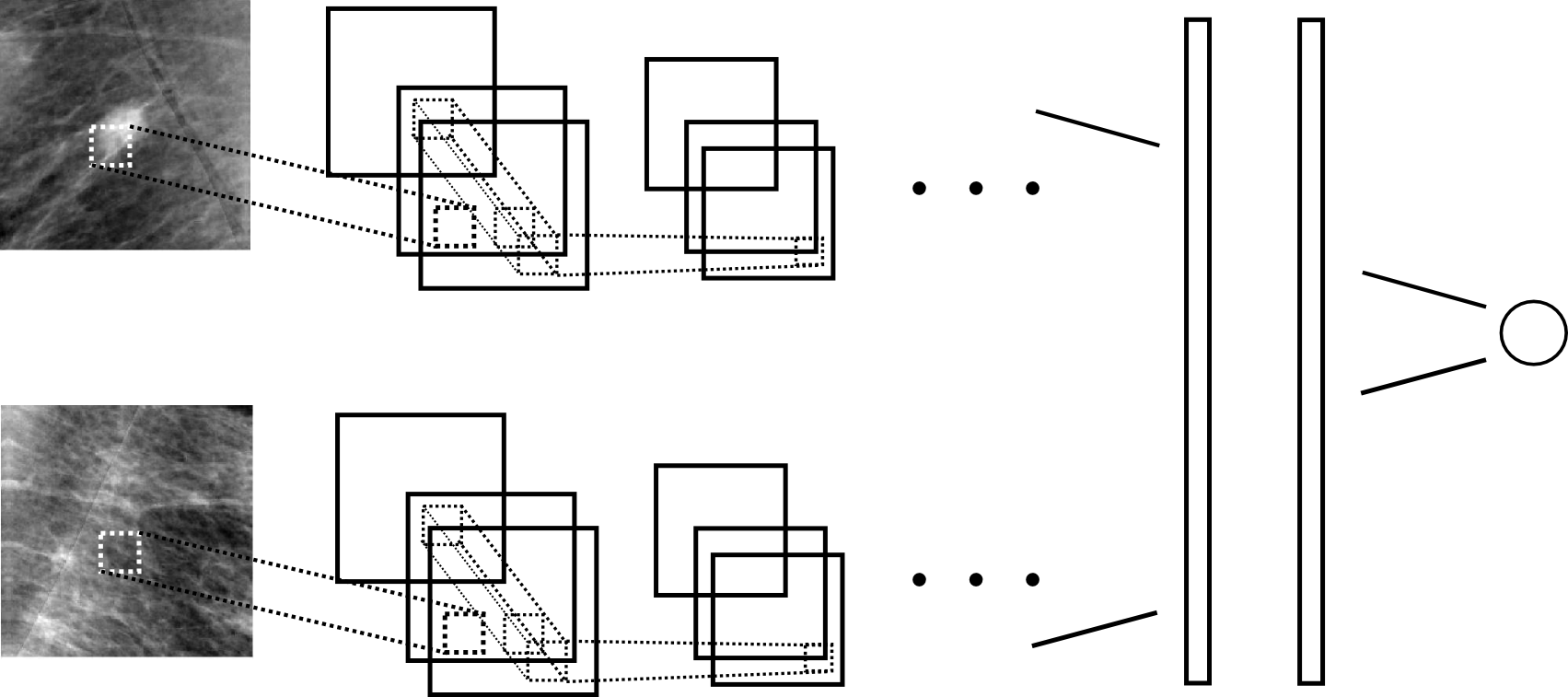}
 \caption{To learn differences between left and right breast and temporal change around a candidate location, we use a two-stream Convolutional Neural Network (CNN). The first stream has as input a patch centered at a candidate location, the second stream a patch around the same location in either the contra-lateral image or the prior, using the mapping depicted in figure \ref{fig::symmetry_mapping_example}. All weights are shared across streams and feature maps are concatenated before the first fully connected layer.}
 \label{fig::network_illustration}
\end{figure}

Partly inspired by the work of Karpathy et al. \cite{Karp14}, we propose to add the contra-lateral and (first prior) temporal counterparts of a patch as separate datastreams to a network. In principle, the datastreams can be merged at any point in the network, with simply treating the additional patch as a second channel the extreme case. Neverova et al. \cite{Neve14} postulate the optimal point of fusion pertains to the degree of similarity of the sources, but to the best of our knowledge no empirical of theoretical work exists that investigates this. We evaluate two architectures:
\begin{enumerate}
 \item A two-stream network where kernels are shared and datastreams are fused at the first fully connected layer. Figure \ref{fig::network_illustration} provides an illustration of this network. 
 \item A single patch, single stream network is used as a feature extractor by classifying all samples in the training and test set and extracting the latent representation of each patch from the first fully connected layer ${\bf x}^{fc1}$ of the network. This feature representation of the primary and either contra-lateral or prior ROI are concatenated and fed to a 'shallow' GBT classifier to generate a new posterior that captures both symmetry or (first prior) temporal information.
\end{enumerate}
The second approach is far easier to train, since it does not entail re-optimizing hyperparameters of a deep model, which is tedious and time consuming. A downside is that the kernels effectively see less data and are therefore potentially less optimal for the task. Additionally, the second setup is more prone to overfitting. We will elaborate on this in the discussion. \\

In general, there are a lot less temporal than symmetry samples because they require two rounds of screening and symmetry samples only one. To compare these architectures, we could simply take a subset of the data where each current exam has both a contra-lateral and prior counterpart. Unfortunately, this yields a relatively small number of positive samples and in early experiments, we found the (base) performance to be very marginal and not sufficient to provide a fair comparison. We therefore view missing prior exams simply as missing data. Although missing data has been well studied in the statistics community \cite{Alli01}, relatively little has been published with respect to discriminative models. \\

In the context of recurrent neural networks (RNNs) \cite{Grav12, Gref15, Lipt15a}, several imputation methods have been explored \cite{Che16, Zach16}. Lipton et al. \cite{Zach16} investigate two imputation strategies: {\it zero-imputation}, where missing samples are simple set to zero and {\it forward-filling} that sets the missing value to the value observed before that. Their results show zero imputation with missing data indicators works best, but no significance analysis is performed. In a similar spirit we explore two strategies:
\begin{enumerate}
 \item use a black image when no prior is available. When a woman skipped a screening round, we map the image to the exam four years before the current or add a black image if this is absent.
 \item use the image from the exam four years before the current image and use the current when no prior is available. 
\end{enumerate}
The first approach carries some additional information, in the sense that the absence of a prior may also increase the likelihood that an exam is positive, since more cancers are typically found in the first round of screening. In the second setting, it is difficult for the network to distinguish pairs where no change is observed and pairs where simply no prior is available.

To add symmetry and temporal information simultaneously, both architectures can trivially be extended with a third stream. However, this requires some additional engineering and we therefore restrict this study to learning two separate models and will propose ways to extend this in the discussion. 

\section{Experiments}
\label{sec::experiments}
\subsection{Data}
Our data was collected from a mammography screening program in The Netherlands (Screening Mid-West) and was recorded with a Hologic Selenia mammography device at an original resolution of 70 micron. All malignant masses were biopsy proven and annotated using contours drawn under the supervision of experienced radiologists. A candidate was considered positive, if the locations was in or within 0.7 cm from an annotated malignant lesion. Before presentation to the human reader, the image is typically processed to optimize contrast and enhance the breast periphery. To prevent information loss, we work on the raw images instead and only apply a log transform which results in a representation in which attenuation and pixel values are linearly related. The images are subsequently scaled to 200 micron using bilinear interpolation.

Our dataset consists of 18366 cases of 18366 women. Each case comprises of one or more exams taken at intervals of two years, unless a women skipped a screening. Each exam again consists of typically four images: a craniocaudal and mediolateral oblique view of each breast. We generated training, validation and test set by splitting on a case level, i.e., samples from the same patient are not scattered across sets. We took $65\%$ for training, $15\%$ for validation and $25\%$ for testing. An overview of the data is provided in table \ref{tab::data_overview}. 

\begin{table}
\centering
 \caption{Overview of the data used for training, validation and testing. Findings refers to the amount of candidates (before data augmentation). Number are separated by '/' where the first number indicates the amount for training, the second the amount for validation and the third the amount for testing.}
 \label{tab::data_overview}
 \begin{tabular}{|c|c|c|c|}\hline
			&    {\bf Findings}		&	{\bf Cases}	\\\hline
  Masses		&	869/210/470		&	796/189/386	\\
  Normal		&	200982/54566/74799	&	3111/1482/1137	\\\hline
 \end{tabular}
\end{table}

\subsection{Learning Settings and Implementation Details}
The networks were implemented in TensorFlow \cite{Abad16} and generally follow the architecture used in Kooi et al. \cite{Kooi16}. Hyperparameters of all models were optimized on a separate validation set using random search \cite{Berg12d}. For the deep CNNs, we employed VGG-like \cite{Simo14} architectures with 5 convolutional layers with $\{16, 16, 32, 32, 64\}$ kernels of size $3 \times 3$ in all layers. We used 'valid' convolutions using a stride of 1 in all settings. Max pooling of $2 \times 2$ was used using a stride of 1 in all but the final convolutional layer. Two fully connected layers of $512$ each were added. Weights were initialized using the MSRA weight filler \cite{He15}, with weight sampled from a truncated normal, all biases were initialized to $0.001$. We employed ELU's \cite{Clev15} as transfer functions in all layers. Learning rate, dropout rate and L2 norm coefficient were optimized per architecture. \\

Since the class ratio is in the order of 1/10000, randomly sampling minibatches will result in very poor performance as the network will just learn to classify all samples as negative. We therefore applied the following scheme. We generated two separate datasets, one for all positive and one for all negative samples. Negative samples are read from disk in chunks and all positive samples are loaded into host RAM. During an epoch, we cycle through all negative samples and in each minibatch take a random selection of an equal amount of positives, which are subsequently fed to GPU where gradients are computed and updated. This way, all negative samples are presented in each epoch and the class balance is maintained. Each configuration trained for roughly 10 days on a TitanX 12 GB GPU. \\

For the shallow model, we employ Gradient Boosted Trees (GBT) \cite{Frie01} using the excellent XGBoost implementation \cite{Chen16d}. We cross-validated the shrinkage and depth using 16 folds. Further parameters were tuned on a fixed validation set using a coordinate descent like scheme. Since the last fully connected layer has size 512, the input to the GBT comprised of 512 features for the single patch setting and a feature vector of 1024 in the symmetry and temporal setting.

\subsection{Results}
Given the results from clinical literature regarding the merit of priors, we focus our results on the classification of candidates and therefore only present ROC curves, rather than FROC curves that are commonly used for detection. To obtain confidence intervals and perform significance testing, we performed bootstrapping \cite{Efro94} using 5000 bootstraps. All curves shown are the mean curve from these bootstrap samples using cubic interpolation. The baseline obtained an AUC of $0.87$ with confidence interval $[0.853, 0.893]$. \\

\begin{figure}
 \centering
 \includegraphics[width=0.5\textwidth]{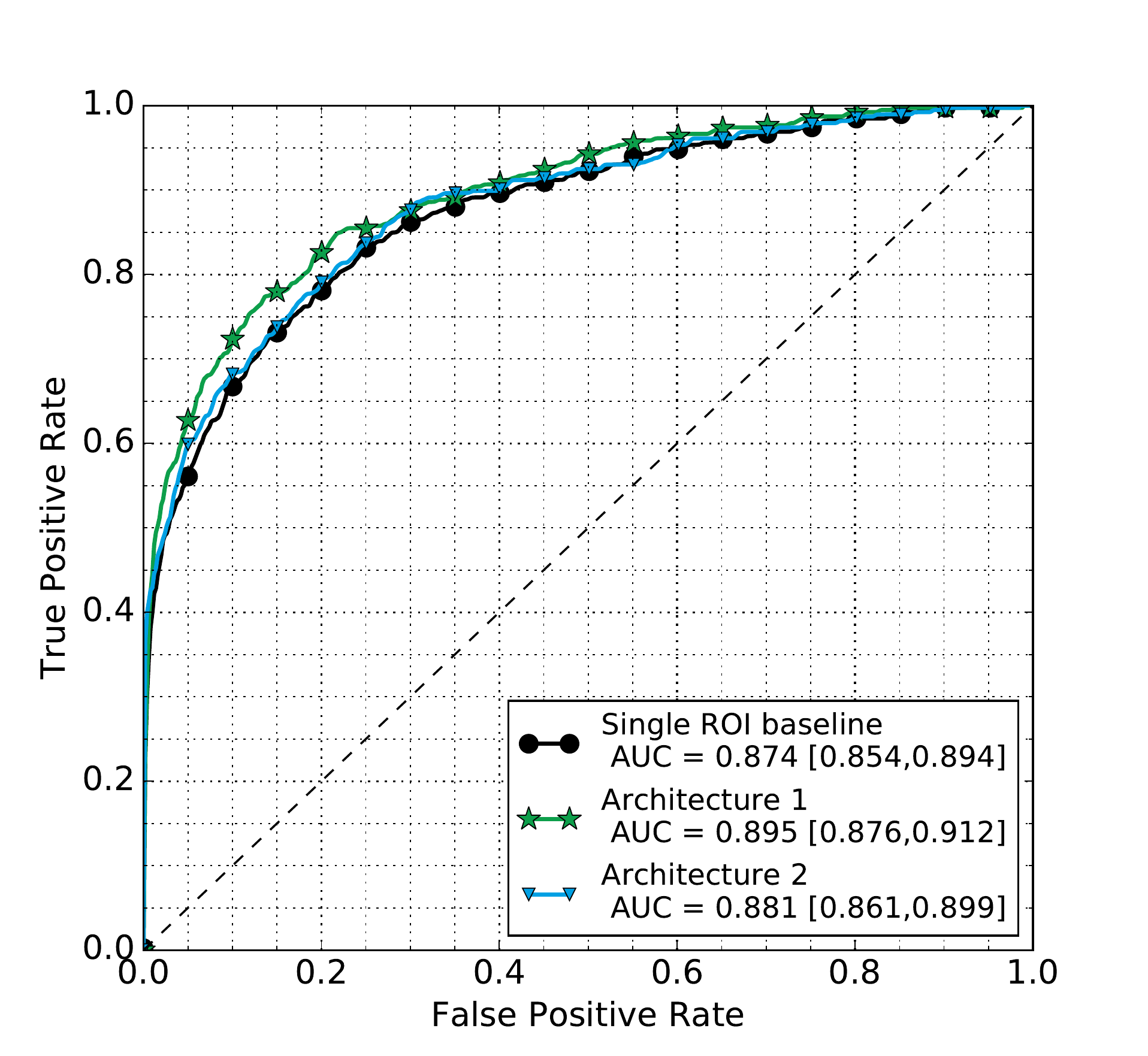}
 \caption{ROC curves of the baseline CNN using a single ROI and the two fusing architectures described in section \ref{sec::fusion_architectures} when presented with the contra-lateral ROI.}
 \label{fig::symmetry_comparison}
\end{figure}

Figure \ref{fig::symmetry_comparison} shows the results of the single ROI baseline, and the fusion architectures as described in section \ref{sec::fusion_architectures} applied to the symmetry comparison. The first architecture where both patches are presented during training obtained an AUC of $0.895$ with confidence interval $[0.877, 0.913]$ and the second architecture where a new classifier is retrained on the concatenation and AUC of $0.88$ with confidence interval $[0.859, 0.9]$. We find significant difference at high specificity on the interval $[0, 0.2]$, $p = 0.02$ between the first architecture and the baseline, but no significant difference on the full AUC ($p = 0.14$). For the second architecture we did not find a significant difference between either the baseline or the first architecture. 

\begin{figure}
 \centering
 \subfloat{\includegraphics[width=0.45\textwidth]{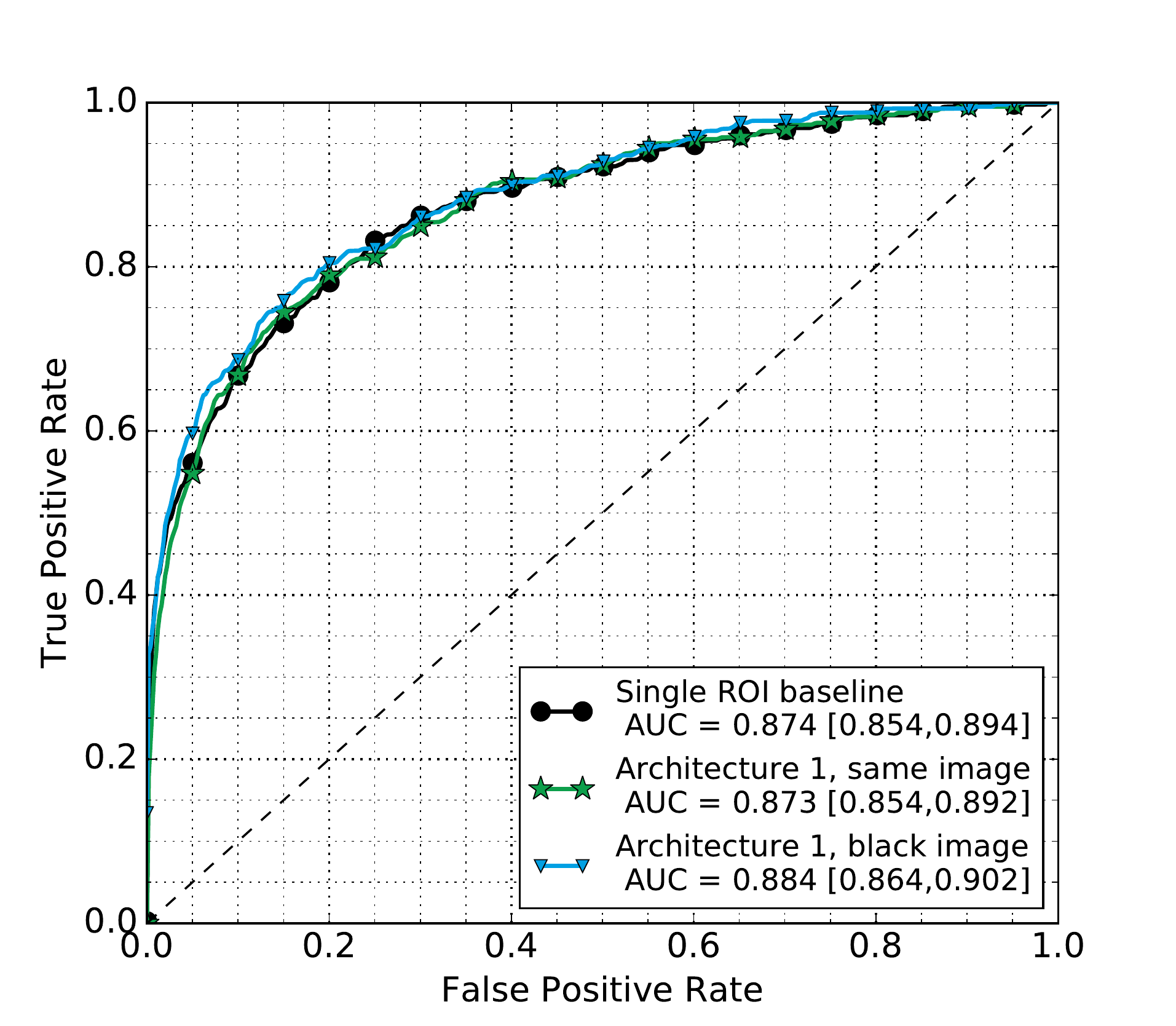}}
 \subfloat{\includegraphics[width=0.45\textwidth]{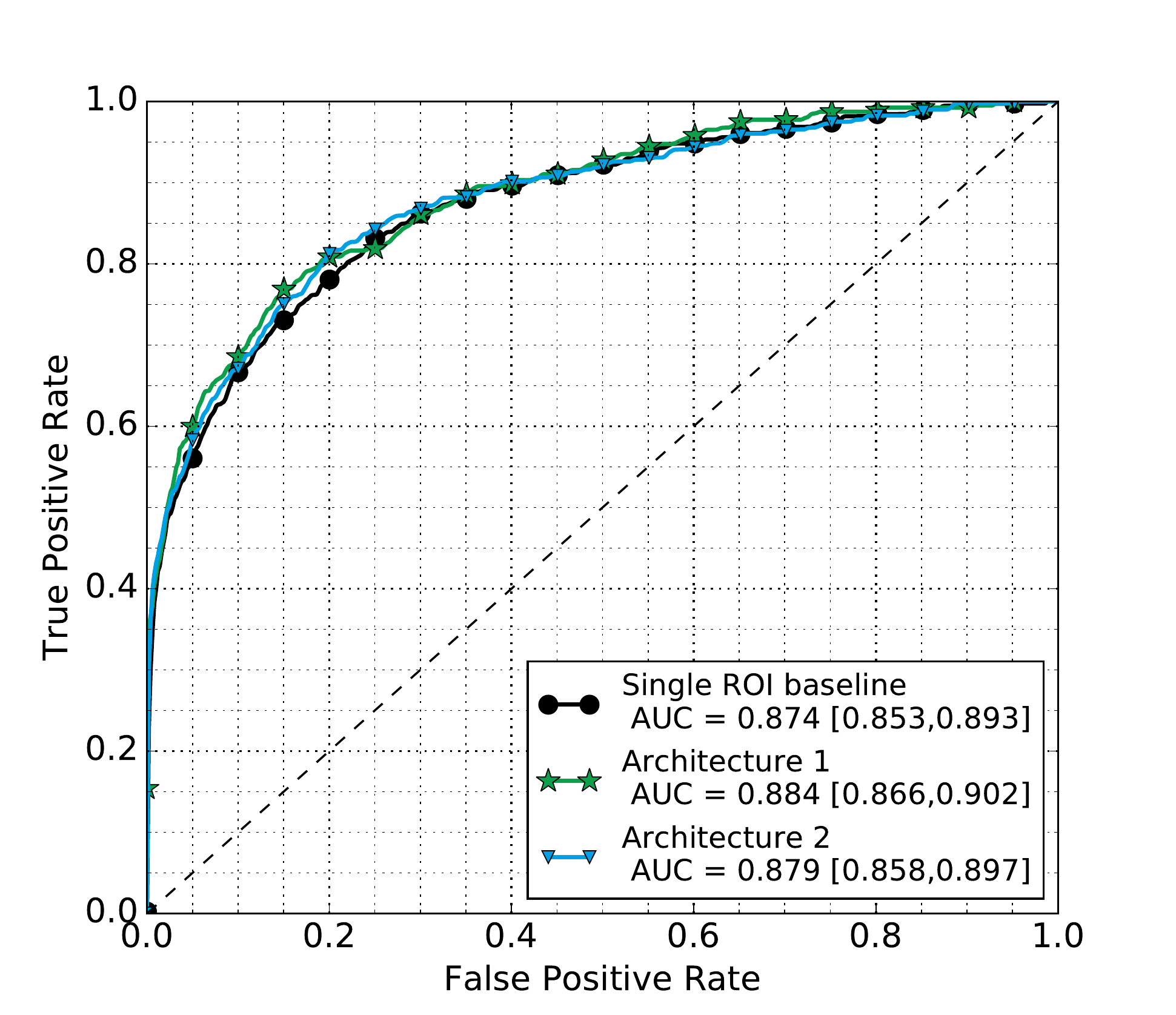}}
 \caption{(a) ROC curves of the baseline CNN using a single ROI and the two strategies to handle missing prior images both using architecture 1. (b) ROC curves of the baseline CNN using a single ROI and the two fusing architectures described in section \ref{sec::fusion_architectures} when presented with the prior ROI and black image strategy.}
 \label{fig::temporal_comparison}
\end{figure}

Figure \ref{fig::temporal_comparison} shows the results of the single ROI baseline and the fusion architectures applied to the temporal comparison. We first investigated the difference between the two different strategies to handle missing priors. The approach using the same image obtained an AUC of $0.873$ with confidence interval $[0.854, 0.892]$, the approach using the black image for missing priors an AUC of $0.884$ with confidence interval $[0.866, 0.902]$. We did not see a significant difference between the strategies $p >> 0.05$, however, the strategy where the black image was used has a higher AUC and we have decided to use this to compare the fusing architectures. \\
The first architecture where both patches are presented during training obtained an AUC of $0.884$ with confidence interval $[0.866, 0.902]$ and the second architecture where a new classifier is retrained on the concatenation an AUC of $0.879$ with confidence interval $[0.858, 0.898]$.  We did not find a significant difference between any of the architectures $p >> 0.05$, but improvements were found to be consistent during early experiments. Results will be discussed in the following section.

\section{Discussion}
\label{sec::discussion}
From the curves in figure \ref{fig::symmetry_comparison} and \ref{fig::temporal_comparison} we can see both symmetry and temporal data improve performance, but only see marginal improvements with temporal data. The curves also show the scheme where both ROIs are fed to a single network (architecture (1) in section \ref{sec::fusion_architectures}) works best. As mentioned in section \ref{sec::fusion_architectures}, architecture (2) has the advantage that no new networks need to be trained which can take several months to do properly for large datasets. Two disadvantages, however, are that (1) the kernels in the network (parameters up to the first fully connected layer) effectively see less data. In the first architecture, even though the kernels are shared, they are trained on both the primary and either symmetry or prior patch and therefore better adjusted to the task. (2) Overfitting is a much bigger issue: since the features are learned on most of the data the models are trained on, the cross-validation procedure of the GBT often gave a strong underestimate of the optimal regularization coefficients (depth, shrinkage in the case of the GBT), resulting in strong gaps between train and test performance. Optimizing this on a fixed validation set did not result in much better performance. We have tried extracting features from deeper in the network to mitigate this effect but found lower performance. \\ 

Since many exams do not have a prior, we explored two strategies to fill in this missing data. In the first setting, we used a black image when no prior image was available and in the second strategy, the same image as the current was used. From the curves in figure \ref{fig::temporal_comparison} we can see that in the first setting the prior ROI does add some information and therefore this approach is at least not detrimental to performance. In the second setting, however, we do not see an increase. A possible advantage of the first approach is that it carries some additional information: the number of tumors found in the first screening round is often higher, when using imputation methods mentioned by Lipton et al. \cite{Lipt15a} this information is effectively lost. As also mentioned in section \ref{sec::fusion_architectures}, the disadvantage of the second approach is that it is difficult for the network to distinguish between malignant mass-no prior pairs and malignant mass-malignant mass pairs, since no change is typically associated with normal tissue. \\

In clinical practice, radiologists sometimes look back two studies instead of one, when comparing the current to the prior. Since this requires three screening rounds, this reduces the size of our dataset again, if we want to emulate this and more prior ROIs need an imputed image. Ideally, the neural network architecture should accommodate a varying set of priors. In early experiments, we have explored the use of Recurrent Neural Networks \cite{Grav12, Gref15, Lipt15a}, a model designed for temporal data that can be trained and tested on varying input and output sizes. We did not see a clear improvement in performance, but plan to explore this idea more in future work. Since this model can work with varying length inputs, it also provides an elegant way to handle missing prior exams.\\

In this study, we have trained all networks from scratch. Since the rudimentary features that are useful to detect cancer in one view are expected to be almost as useful when combining views, a better strategy may be to initialize the symmetry or temporal two-stream network with the weights trained on a single ROI. Similarly, since we expect similar features are useful to spot discrepancies between left and right breast as to spot differences between time points, the temporal network could be initialized with the network trained on symmetry patches or the other way around. Due to time constraints this was left to future work, but we suspect an increase in performance. \\

We have compared two different fusion strategies. As mentioned in section \ref{sec::fusion_architectures}, the datastreams can in principle be fused at any point in the network, as done by Karpathy et al. \cite{Karp14}. However, there is no guarantee that different architectures perform optimal using the same hyperparameters. For instance, the weight updates of lower layers change if fusion is performed at different points higher in the network. In particular, the learning rate is often found to be important and we feel comparison rings somewhat hollow if no extensive search through the parameter space is done. Since a model typically trains for roughly a week, this is infeasible with our current hardware and we have decided to focus on the two presented models. \\ 

Since the focus of this paper is the presentation of two fusion schemes for adding symmetry and temporal information to a deep CNN, we have presented separate results for each. In practice, when using a CAD system to generate a label for a case, these should be merged into one decision. As mentioned in section \ref{sec::fusion_architectures}, extending the network with a third datastream is trivial. However, this limits the application to cases where both prior and contra-lateral image are available. In our method, we have added a black image, where priors where not available and a similar approach could be pursued in this setting. Another option would be to train a third classifier on top of the latent representation from separate CNNs or the posterior output by separate CNNs, possibly using a missing data model. Since training deep neural networks and optimizing hyperparameters takes a lot of time, we have left this for future work. 
\section{Conclusion}
\label{sec::conclusion}
In this paper we have presented two deep Convolutional Neural Network (CNN) architectures to add symmetry and temporal information to a Computer Aided Detection (CAD) system for mass candidates in mammography. To the best of our knowledge, this is the first approach exploring deep CNNs for symmetry and temporal classification in a CAD system. Results show improvement in performance for both symmetry and temporal data. Though in the latter case gain in performance is still marginal, it is promising and we suspect that when more data becomes available, performance will significantly increase. Although the methods are applied to mammography, we think results can be relevant for other CAD problems were symmetrical differences within or between organs are sought, such as lung, brain and prostate images or CAD tasks where temporal change needs to be analyzed, such as lung cancer screening. \\

\section*{Disclosures}
Thijs Kooi has no potential conflicts of interest. Nico Karssemeijer is co-founder, shareholder, and director of ScreenPoint Medical BV (Nijmegen, The Netherlands), co-founder of Volpara Health Technologies Ltd. (Wellington, New Zealand), and QView Medical Inc. (Los Altos, CA).

\section*{Acknowledgements}
This research was funded by grant KUN 2012-5577 of the Dutch Cancer Society and supported by the Foundation of Population Screening Mid West.


\end{document}